\documentclass{ecai} 

\usepackage{latexsym}
\usepackage{amssymb}
\usepackage{amsmath}
\usepackage{amsthm}
\usepackage{booktabs}
\usepackage{enumitem}
\usepackage{graphicx}
\usepackage{color}

\usepackage{algorithm}

\usepackage{graphicx}
\usepackage{textcomp}
\usepackage{xcolor}
\usepackage{amsmath}
\usepackage{adjustbox}
\usepackage{booktabs}
\usepackage{algpseudocode}

\newcommand{\BibTeX}{B\kern-.05em{\sc i\kern-.025em b}\kern-.08em\TeX}

\begin{document}


\begin{frontmatter}


\paperid{123} 


\title{Semantic Adapter for Universal Text Embeddings: Diagnosing and Mitigating Negation Blindness to Enhance Universality}


\author[A]{\fnms{Hongliu}~\snm{CAO}\orcid{0000-0002-1326-8159}\thanks{Corresponding Author. Email: hongliu.cao@amadeus.com.}
}

\address[A]{Amadeus France}


\begin{abstract}

Text embeddings are crucial for various natural language processing tasks, gaining popularity in both industry and academia. Recent progress in training data and LLMs has greatly enhanced the development of universal text embeddings, aiming for a single unified model that can handle diverse tasks, domains and languages. However, due to biases inherent in popular evaluation benchmarks, certain perspectives or abilities of these models remain unassessed. One such overlooked aspect is the models' capacity for negation awareness.
To address this gap in the existing literature, this paper presents a comprehensive analysis of the negation awareness in state-of-the-art universal text embedding models. Our investigation reveals a substantial deficiency in these models, as they frequently interpret negated text pairs as semantically similar. This flaw undermines their effectiveness in accurately understanding negated statements.
To mitigate this issue and enhance the universality of text embeddings, we introduce a lightweight, parameter-free negation adapter. The proposed solution involves a data-efficient and computational-efficient embedding re-weighting method that does not require modifications to the parameters of existing text embedding models. The proposed solution is able to improve text embedding models' negation awareness significantly on both simple negation understanding task and complex negation understanding task. Furthermore, the proposed solution can also significantly improve the negation awareness of Large Language Model based task-specific high dimensional universal text embeddings.
These results not only bridge a key “universality gap” but also pave the way for a modular semantic adapter paradigm toward more universal, robust, and environmentally conscious text embeddings.

\end{abstract}
\end{frontmatter}

\section{Introduction}

Text embedding has gained significant attention from both industry and academia due to its crucial role in various Natural Language Processing (NLP) tasks such as information retrieval \cite{rajapakse2023dense}, sentiment analysis \cite{suresh2021not,zhang2022leveraging}, text clustering \cite{xu2023contrastive,sBERTreimers2019sentence},  etc.
Recently, the importance of text embeddings has been further highlighted in the context of Large Language Models (LLMs) based applications, such as Retrieval Augmented Generation (RAG) systems. This is primarily because these applications rely on high-quality text embeddings for performing vector search, which involves retrieving the most relevant contexts/documents for LLM Question Answering (QA) tasks \cite{cao2024recent}. 

The field of text embeddings has recently achieved significant advancements, transitioning from contextual embeddings like BERT \cite{devlin2018bert} to universal text embeddings. These universal models aim to create a comprehensive embedding framework capable of handling varied input lengths, downstream tasks, domains, and languages. This progress can be attributed to the advances in the quantity and quality of diverse text datasets across different tasks \cite{bgexiao2023c,asai2022task}, synthetic data generated by different LLMs \cite{lee2024gecko,e5mistralwang2023improving}, using LLMs as backbones and the emergence of benchmarks with the focus on novel task and domain generalization such as the Massive Text Embedding Benchmark (MTEB) \cite{muennighoff2022mteb}.  Representative universal embedding models such as  GTE \cite{gteli2023towards}, BGE \cite{bgexiao2023c}, E5 \cite{e5wang2022text,e5mistralwang2023improving,e5instructwang2024multilingual}, Gecko \cite{lee2024gecko} or LLM2Vec \cite{llm2vec} can effectively address diverse downstream tasks without fine-tuning. They have demonstrated significant improvements over contextual text embeddings across diverse tasks including information retrieval, reranking, clustering and pair classification tasks \cite{cao2024recent}.  However, evaluation benchmarks such as MTEB are biased towards tasks such as retrieval, classification and clustering \cite{muennighoff2022mteb, cao2024recent, cao2024writing}. 

Despite the widespread adoption of popular evaluation benchmarks, these frameworks often fail to assess certain critical capabilities of text embedding models. 
In particular, the negation awareness remains underexamined: a fundamental linguistic construct that plays a pivotal role in understanding sentiment and meaning \cite{cannotanschutz2023not}.
Its misrepresentation can alter meaning, distort sentiment analysis, and invalidate inference in applications such as opinion mining and fact verification \cite{cannotanschutz2023not, ribeiro2020beyond, hossain2022analysis}. 
Previous works have shown that contextual text embeddings like BERT \cite{devlin2018bert} lack understanding of negations and fail to attribute sufficient importance to the word “not” \cite{ettinger2020BERTnot, hossain2020analysis, hossain2022analysis}. This bias can compromise the model's output, thereby undermining the reliability of these systems, particularly in scenarios involving opinion mining, fact-checking, or assessing answer correctness \cite{cannotanschutz2023not}. However, the extent of this “negation blindness” in universal embeddings has not yet been systematically studied.

This work bridges the gap in existing literature and  makes the following contributions:
\begin{itemize}
    \item We formalize negation as a semantic stress test for universal embeddings, constructing controlled paraphrase–negation contrast sets to examine negation awareness across embedding models.
    \item  We introduce the negation adapter, a data and compute efficient re-weighting module that amplifies negation-sensitive dimensions without any fine-tuning of the embedding models.
    \item We demonstrate that the proposed solution is able to improve text embedding models' negation awareness significantly on both simple negation understanding task and complex negation understanding task. 
    \item We further demonstrate that  the proposed solution can also significantly improve the negation awareness of LLM-based task-specific high dimensional universal text embeddings.   
\end{itemize}

\section{Related Works}
\subsection{Universal text embeddings}
The field of text embeddings has seen substantial improvements in recent years. Text embedding models can be categorized in four eras  \cite{cao2024recent}: the first era stands for count-based embeddings such as Bag of Words \cite{harris1954distributional}  and Term Frequency-Inverse Document Frequency (TF-IDF) \cite{manning2008introduction} which are transformed into low-dimensional dense embeddings with methods like Latent Semantic Indexing (LSA) \cite{deerwester1990indexing}; the second era is static dense word embeddings represented by Word2Vec \cite{word2vec}, GloVe \cite{glove} and FastText \cite{fasttext} which can not deal with polysemy; the third era is contextualized embeddings where context-sensitive dynamic embeddings can adapt or change based on context and deal with polysemy, represented by Embeddings From Language Models (ELMo) \cite{elmoneumann2018deep}, Generative Pre-trained Transformer (GPT) \cite{gpt1radford2018improving} and Bidirectional Encoder Representations from Transformers (BERT) \cite{devlin2018bert}. Recently, there has been significant progress in the fourth generation of universal text embeddings which aim at building a unified comprehensive text embedding to address a multitude of input text length, tasks, domains and languages \cite{cao2024recent}. 

To achieve universality, large quantity datasets with diverse mixtures are used for both pre-training and fine-tuning by GTE \cite{gteli2023towards}. BGE \cite{bgexiao2023c} focuses on improving the training data quality by filtering irrelevant text pairs and using multi-task high quality labeled data for fine-tuning. Similarly,  E5 \cite{e5wang2022text} also focuses on data quality improvement by constructing a curated web-scale text pair dataset  named Colossal Clean text Pairs (CCPairs), while Multilingual E5 \cite{e5instructwang2024multilingual} combines diverse real world datasets with synthetic datasets containing 150k unique instructions and 93 languages generated by GPT-3.5/4 in order go generalize across different tasks as well as different languages. Some other universal text embeddings focus on new loss functions including Universal AnglE Embedding (UAE) \cite{li2023angle} and mxbai-embed-large-v1 from \cite{mrlkusupati2022matryoshka, 2dmrlli20242d}.  LLMs are also used as backbones for universal text embeddings (e.g. E5-Mistral \cite{e5mistralwang2023improving}, LLM2Vec \cite{llm2vec}, gte-Qwen1.5-7B-instruct \cite{gteli2023towards}) as they do not need the large-scale contrastive pre-training.

\subsection{Negation understanding}

\begin{figure}[h]
  \centering
  \includegraphics[width=0.7\linewidth]{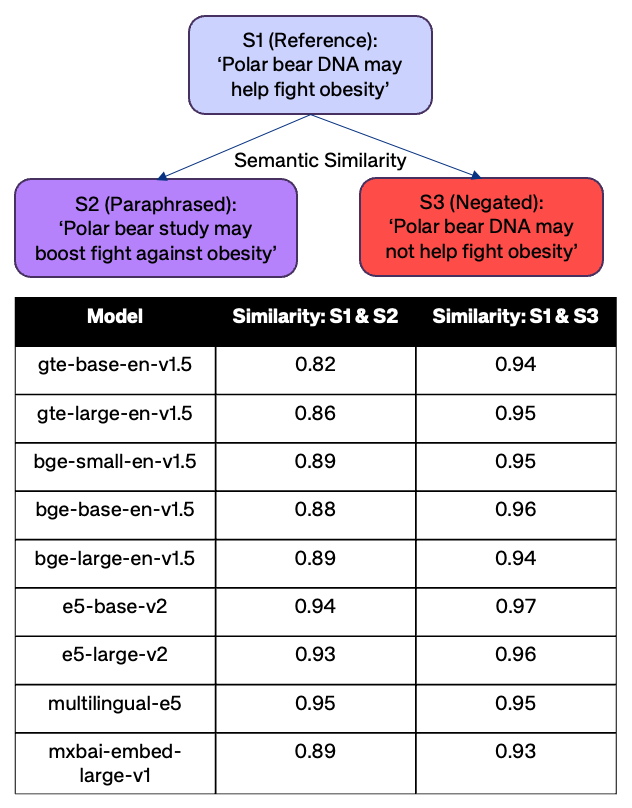}
  \caption{Example of semantic similarity based on universal text embeddings. S1 and S2 are paraphrases from STSB \cite{stsbcer2017semeval} with human annotated similarity of 0.96. S3 is a simple negation of the reference S1. The cosine similarity between text embeddings of S1 and S2 as well as S1 and S3 are compared.} 
  \label{fig:eg}
\end{figure}

Negation plays an important role in many NLP tasks including Natural Language Inference (NLI) \cite{cannotanschutz2023not}, sentiment analysis  \cite{ribeiro2020beyond}, Natural Language Understanding (NLU) tasks \cite{hossain2022analysis}, etc. 
The complexity of negation, as highlighted by \cite{ribeiro2020beyond}, presents substantial challenges for commercial sentiment analysis.
  Contextual text embedding models such as BERT \cite{devlin2018bert}  is shown to be insensitive to negations in \cite{ettinger2020BERTnot} through a completion task involving masked hypernyms in affirmative and negated versions of each sentence. 
  Other Contextual text embeddings including Transformer-XL \cite{dai2019transformer} and ELMO \cite{elmoneumann2018deep} are also found  to  exhibit poor performance in distinguishing between affirmative and negative sentences when subjected to negative cloze tests \cite{kassner2019negated2}. Moreover, they tend to make identical predictions regardless of the polarity of the sentence \cite{vahtola2022not}.  
In the study conducted by \cite{hossain2020analysis, hossain2022analysis},  a novel benchmark was created to evaluate the ability of contextual embedding models in terms of recognizing negation. The experimental findings revealed that even after fine-tuning with additional negation-containing pairs, contextual text embeddings (RoBERTa \cite{liu2019roBERTa}, XLNet \cite{yang2019xlnet} and BERT \cite{devlin2018bert}) still face challenges in accurately understanding negation.

Several studies have been conducted in order to improve the negation awareness of contextual text embedding models such as BERT. In \cite{hosseini2021understanding}, BERTNOT is proposed by fine-tuning a pre-trained BERT model using  a knowledge distillation objective as well as a novel unlikelihood objective, which is based on negated generic sentences extracted from a raw text corpus. The proposed BERTNOT  has been observed to effectively incorporate negation knowledge into the BERT-base model. However, it has been found to be ineffective with the BERT-large model. 
A novel negation-focused pre-training strategy is proposed in \cite{truong2022improving} to improve the performance and generalization ability of negation detection using  targeted data augmentation and negation masking. Another new pre-training framework with a weighted cross-entropy loss and elastic weight consolidation regularization was proposed in \cite{singh2023nlms}
to overcome catastrophic forgetting and augment the negation knowledge in contextual  embeddings.

Compared to contextual embeddings, universal embeddings have the capacity of generalizing across diverse tasks.
However,  their negation awareness is yet to be determined as the evaluation benchmarks such as MTEB are biased towards tasks with many datasets, notably retrieval, classification and clustering \cite{muennighoff2022mteb, cao2024recent}.
A simple example is shown in Figure \ref{fig:eg}: given the reference sentence S1 ('Polar bear DNA may help fight obesity') and a semantic similar sentence S2 ('Polar bear study may boost fight against obesity') from Semantic Textual Similarity Benchmark (STSB) \cite{stsbcer2017semeval} with the groudtruth similarity value 0.96, S3 is the negated version of S1. Diverse universal text embeddings are used to measure the semantic similarity between S1 \& S2 and the semantic similarity between S1 \& S3. While the semantic similarity between S1 \& S2 is expected to be much larger than the one between S1 \& S3, most universal text embeddings fail with this simple example. 
To understand better the negation awareness of the state-of-the-art universal text embeddings, holistic analysis and comparisons are conducted in the following sections.

\section{Universality Gap Diagnosis}

Semantic similarity refers to the degree of overlap in meaning  \cite{stsbcer2017semeval}.
 Measuring the semantic similarity using cosine over text embeddings  is one of the most adopted method \cite{muennighoff2022mteb}. 
Text embedding based semantic similarity is also used as evaluation metrics to measure the answer correctness of RAGs or LLMs \cite{es2023ragas,adlakha2024evaluating}. 
Despite its  wide usage, there is few study trying to understand what kind of semantic information is encoded and highlighted in the semantic similarity based on text embeddings. 
 In the literature of Semantic Textual Similarity (STS) studies such as \cite{agirre2013sem, stsbcer2017semeval}, semantic similarity is defined as (re-scaled to the range between 0 and 1): 
 \begin{itemize}
     \item 0 means the pair of texts are on different topics;
     \item 0.2 means the pair of texts are not equivalent, but are on the same topic;
     \item 0.4 means  the pair of texts are not equivalent, but share some details;
     \item 0.6 means  the pair of texts are roughly equivalent, but some important information differs/missing;
     \item 0.8 means  the pair of texts are mostly equivalent, but some unimportant details differ; 
     \item 1 means  the pair of texts are completely equivalent;
 \end{itemize}
 
One limitation of this definition is that: the semantic similarity is mainly based on topic information while other semantic information such as negation is not included. 
When it comes to two pieces of text where one is the negated version of the other, the situation becomes more complex. For instance, consider these two sentences: (1) "The horse is white." and (2) "The horse is not white.":
from a purely syntactic perspective, the sentences are very similar. They share the same subject and predicate, with the only difference being the presence of negation in (2). From a purely topic perspective, the sentences are similar as they share similar topics: the color of the horse.  However, semantic textual similarity is not just about the structure or the words used, but about the meaning conveyed by the sentences. The negation in the second sentence changes the meaning entirely.
According to \cite{singh2023nlms},  the significance of negation in understanding the semantic relationship in natural language texts is profound, as it reverses the overall semantic meaning of a sentence. Consequently, we formalize negation as a stress test of embedding universality and robustness in this section to study how universal text embeddings interpret the semantic similarity between negated text pairs.

\subsection{Dataset}

STSB \cite{stsbcer2017semeval} is a collection of English datasets that have been used in the SemEval and *SEM STS shared tasks over the period from 2012 to 2017 \cite{stsbcer2017semeval}. 
Annotation of text pair's similarity is accomplished via crowdsourcing, integrating pragmatic and global knowledge, thereby enhancing their interpretability and utility for subsequent tasks \cite{stsbcer2017semeval}. Within the scope of this study, the human annotation similarity scores in the STSB dataset, which typically range from 0 to 5, are re-scaled to the range between 0 and 1.
Given STSB text pairs S1 ($sentence\_1$) and S2 ($sentence\_2$) with the human annotated semantic similarity in the form of  [$S1_i$, $S2_i$, $Sim_i$] (where $i$ is from 1 to N), a sentence-level negation tool (focuses on verbal negations and supports the simple addition and deletion of negation cues) developed by \cite{cannotanschutz2023not} is used to create $NegS1_i$: the simple negated version of $S1_i$  (e.g. transform "The horse is white" into its simple negated version of "The horse is not white"). 
The training partition of the dataset is used for analysis (in this section) and training (in next section), while the development and test sets are combined for testing (in next section). 

\subsection{Universal text embeddings}
In this study, diverse universal text embeddings with different parameters size,  training data,  loss functions and  fine-tuning strategies are selected. The focus has been on those that are open-source and have demonstrated superior performance on the MTEB benchmark, including: gte-large-en-v1.5  (434M) and  gte-base-en-v1.5 (137M) \cite{gteli2023towards}, bge-large-en-v1.5  (335M) and bge-base-en-v1.5 (109M) \cite{bgexiao2023c}, e5-large-v2 (335M) and e5-base-v2 with (109M) \cite{e5wang2022text}, multilingual-e5-large-instruct \cite{e5instructwang2024multilingual}  (560M) and  mxbai-embed-large-v1 \cite{mrlkusupati2022matryoshka, 2dmrlli20242d}  (335M).
Two widely used contextual text embeddings from SentenceTransformers \cite{sBERTreimers2019sentence} are selected as baseline models, including all-mpnet-base-v2 (110M) and all-roBERTa-large-v1 (355M).

\subsection{Experiments}

\begin{figure*}[!ht]
  \centering
  \includegraphics[width=\linewidth]{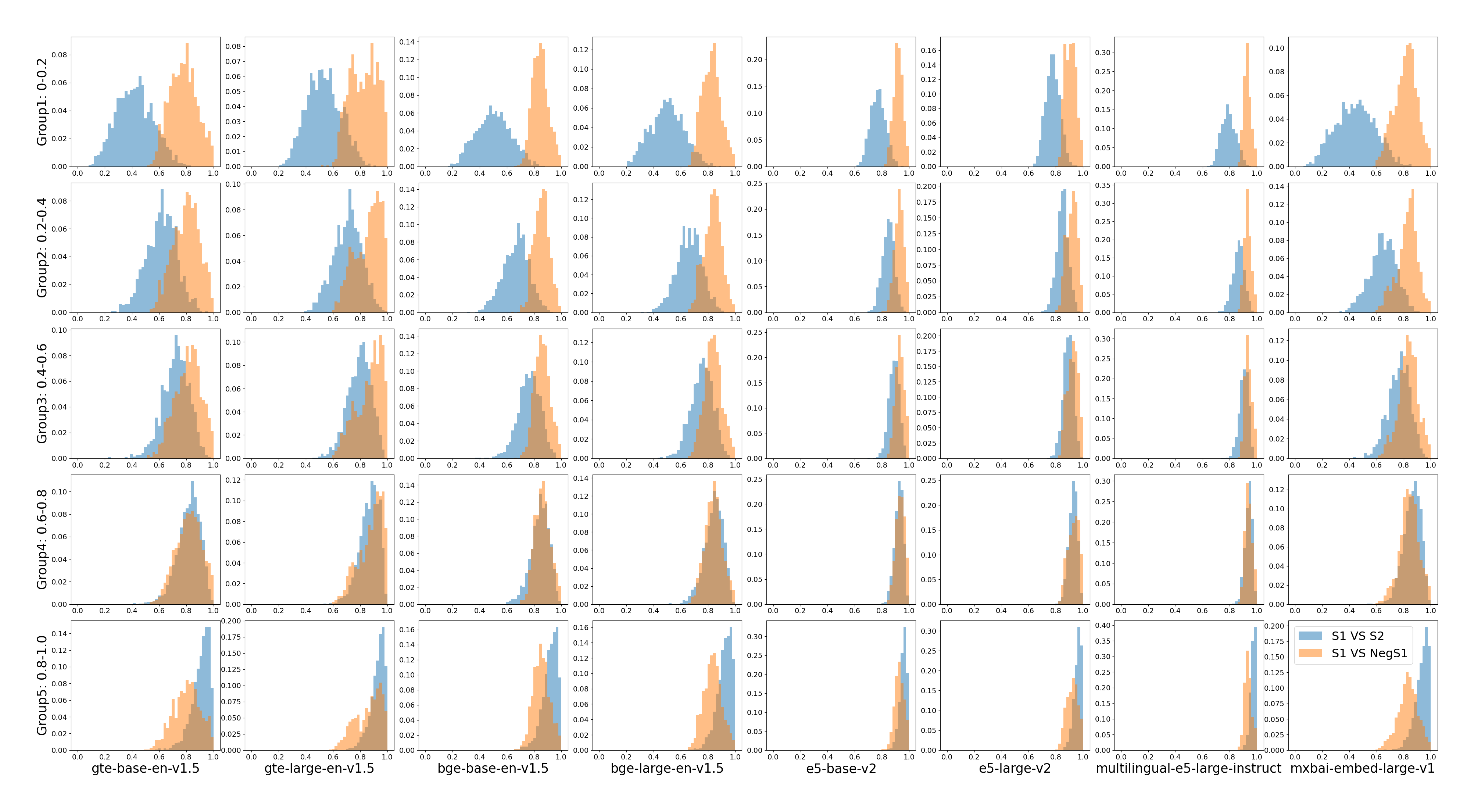}
  \caption{The histogram plot of semantic similarity scores based on universal text embeddings (columns): row 1 to row 5 represents the results of group 1 to group 5. In each subplot, the x-axis is the similarity values between 0 and 1, while y axis is the sample size (in percentage). Blue color represents the semantic similarity between $sentence\_1_i$ and $sentence\_2_i$; orange color represents the semantic similarity between $sentence\_1_i$ and $neg\_sentence\_1_i$. } 
  \label{fig:3}
\end{figure*}

In order to study how universal text embeddings interpret negation, the training data of STSB are separated into five groups based on the human annotated similarity $Sim_i$ between $S1_i$ and $S2_i$: group 1 (0-0.2), group 2 (0.2-0.4), group 3 (0.4-0.6), group 4 (0.6-0.8), and group 5 (0.8-1). 
The semantic similarity $Sim\_12$ between S1  and S2  as well as the semantic similarity $Sim\_1neg1$ between  S1 and its negated version NegS1 within each group based on each universal text embeddings are calculated:

\begin{equation}
    \begin{split}
        Sim\_12_i = Cos( \textbf{e}(S1_i), \textbf{e}(S2_i)) \\
        Sim\_1neg1_i = Cos(\textbf{e}(S1_i), \textbf{e}(NegS1_i))
    \end{split}
\end{equation}

where $\textbf{e}()$ is the embedding model.
By comparing $Sim\_1neg1$  with  $Sim\_12$ relatively within each group, we can get the insights on how similar are negated text pairs interpreted by the state-of-the-art universal text embeddings. 

The histogram of $Sim\_12$ within each group is illustrated in blue color in Figure \ref{fig:3}: row 1 to row 5 represents the results from group 1 to group 5 and each column represents a text embedding model. Generally speaking, the absolute similarity values do not match the human annotation. For instance, the human annotation for group 1 is between 0 and 0.2, while the similarity values calculated by most embedding models  are between 0.2 and 0.8. For E5 series, similarity values  are between 0.6 and 1 for group 1. This substantial discrepancy highlights that the raw cosine similarity scores generated by these models lack interpretability, as they are not calibrated to scales of human-judged similarity.
Consequently, a cosine value of 0.5 cannot be interpreted as "moderately similar" across different models, as each model's absolute scale is arbitrary.
 On the other hand, the evaluation metric used by STSB is  Pearson correlation between embedding based semantic similarity scores and human annotations \cite{stsbcer2017semeval}. Hence, the relative position should also be compared. For all universal text embeddings, the histogram of $Sim\_12$ moves towards the right (more similar) from  data in group 1 (dissimilar text pairs) to data in group 5 (similar text pairs). This result indicates that the universal text embeddings based semantic similarity is correlated with the human annotations.

The histogram of $Sim\_1neg1$  within each group is illustrated as orange color in Figure \ref{fig:3}: the semantic similarity values are within the range [0.6, 1] for all groups and all universal text embeddings. In order to answer how similar are negated text pairs interpreted by the state-of-the-art universal text embeddings, the histogram of $Sim\_1neg1$ is compared with $Sim\_12$ within in each group. For Group 1: The human annotated similarity between S1  and S2  is within [0, 0.2], which means they are very dissimilar. From the first row of Figure \ref{fig:3}, it can be seen that  S1 is more similar to its negated version NegS1 (orange color) than to S2 (blue color) across all universal text embeddings (in 99.27\% cases on average). Similar observations can be found for Group 2 and 3. 
For Group 4: The human annotated similarity between S1 and S2 is within [0.6, 0.8] which means they are roughly similar. From the fourth row of Figure \ref{fig:3}, it can be seen that  the similarity distribution of $Sim\_1neg1$  overlaps with $Sim\_12$.  On average (across all embedding models), S1 is more similar to NegS1 than S2 in 51.57\% cases.  In Group 5: The human annotated similarity between S1 and S2 is within [0.8, 1] which means they are very similar. From the last row of Figure \ref{fig:3}, it can be seen that S1 is more similar to S2 than NegS1 (in 77.25\% cases). 
\textbf{These results reveal a pronounced universality gap: the distribution of $Sim\_12$ and $Sim\_1neg1$ overlaps most within Group 4, suggesting that these embeddings interpreting negated text pairs as roughly similar in terms of semantic meaning.}

\section{Negation Adapter: A Lightweight Alternative to Fine-Tuning}

Our universality gap diagnosis (Section 3) demonstrates that state-of-the-art universal embeddings exhibit “negation blindness”,  underscoring the need for interventions to enhance negation sensitivity for downstream tasks that rely on the semantic understanding of negation.
Previous studies on contexual text embeddings \cite{hosseini2021understanding,truong2022improving, singh2023nlms} assume that these embedding models lack of negation knowledge and  focus on improving negation awareness by  fine-tuning pre-trained models on negation datasets.  However, fine-tuning a universal embedding model on a negation-specific dataset runs counter to the very definition and goals of “universal” text embeddings. In particular:
\begin{itemize}
    \item Universality vs Specialization: a universal embedding aims to perform robustly across diverse input lengths, tasks, domains etc. Fine-tuning on negation data may distort representations for other semantic features (e.g. topics, syntax, etc.), undermining its general-purpose utility and risking performance degradation on other axes.
    \item Fine-tuning large backbones (often hundreds of millions  or billions of parameters) on specialized data is time and energy intensive. This repeated retraining for each new identified universality gap undermines sustainability and eco-friendly deployment. Additionally, each fine-tuned variant necessitates storage, versioning, and integration into pipelines with increasing engineering overhead.
    \item High-quality large size negation datasets are rare and expensive to construct for fine-tuning tasks.
\end{itemize}

To deal with these limitations, we propose a lightweight, post-hoc re-weighting “adapter" which leaves the base model intact and only adjusts its use when negation sensitivity is required. 
The main hypothesis is that universal text embeddings inherently encapsulate a broad but imbalanced spectrum of semantic information including topics, sentiment, negation and so on due to their large scale training/fine-tuning on extensive and diverse datasets. By assigning greater weights to the dimensions that predominantly capture negation-related information, they are expected to recognize and process negations more effectively. \textbf{ To ensure the universality across different tasks: this negation-aware weight will only be applied for negation related tasks, while the original unweighted embedding is used for other tasks.}

Given the D-dimensional embedding $E_{T_i} = \{ e_{T_i}^0, ...  e_{T_i}^D \} $ of text $T_i$, $E_{P_i}$ is the embedding of $P_i$ (the paraphrased version of $T_i$), $E_{N_i}$ is the embedding of $N_i$ (the negated version of $T_i$).  The cosine similarity between $T_i$ and $P_i$ is:
\begin{equation}
\label{cosp}
\begin{split}
    Cos(T_i, P_i) &= \frac{e_{T_i}^1 \times e_{P_i}^1 + ... + e_{T_i}^k \times e_{P_i}^k  + ... + e_{T_i}^D \times e_{P_i}^D     }{ ||E_{T_i}|| \times ||E_{P_i}|| } \\
    & = u_{(T_i, P_i)}^1 + ... + u_{(T_i, P_i)}^k +... + u_{(T_i, P_i)}^D \\
    & where: \quad u_{(T_i, P_i)}^k = \frac{e_{T_i}^k \times e_{P_i}^k}{||E_{T_i}|| \times ||E_{P_i}||}
\end{split}   
\end{equation}

Given the triplet $[T_i, P_i, N_i]$, ideally we want the similarity between the paraphrased text pair $Cos(T_i, P_i)$ to be greater than the similarity between the negated text pair $ Cos(T_i, N_i)$:

\begin{equation}
\label{cospn}
    {Diff}_i = Cos(T_i, P_i) - Cos(T_i, N_i) = \Sigma_{k=1}^D v_{i}^k >0
\end{equation}

where: $ v_{i}^k =  u_{(T_i, P_i)}^k - u_{(T_i, N_i)}^k $. If $ v_{i}^k$ is larger, the dimension $k$ contributes more to the negation awareness, hence the dimension $k$ should be assigned with a larger weight. Given training data with $N$ triplets, the mean contribution of each dimension $k$ can be calculated as:
\begin{equation}
\label{cosm}
   \overline{v}^k = \frac{1}{N} \times \Sigma_{i=1}^N v_{i}^k 
\end{equation}

Given the average contribution of each dimension $\overline{V} = [\overline{v}^1, ... \overline{v}^D ]$ (re-scaled by dividing $max(\overline{V})$ ), the proposed weighting function is:
\begin{equation}
\label{wi}
  w^k = \frac{e^{a\times \overline{v}^k}}{\Sigma_{i=1}^{D}e^{a\times \overline{v}^i } }
\end{equation}
 where $a$ is a hyperparameter to control the trade off between negation information and general semantic information in text embeddings.  Larger $a$ assigns larger weights on to the dimension with positive contribution to negation awareness.
 For negation related tasks, the dimension weights are then applied through element-wise multiplication to the original embeddings. \textbf{This is equivalent to apply a weighted cosine similarity on the original embeddings.}

\section{Experimental results}
\subsection{Experiments on simple negation}
\subsubsection{Datasets} 

In order to test the effectiveness of the proposed embedding re-weighting method, a paraphrase detection task is formulated using the STSB dataset: the Group 5 data from STSB [$sentence\_1_i$, $sentence\_2_i$] are used as paraphrase text pairs.  The  embedding models are then required to identify the accurate paraphrase of $sentence\_1_i$ from [$sentence\_2_i$, $neg\_sentence\_1_i$].
The training split of STSB is used as training data, the validation split and test split of STSB are combined as test data in this work.  

\subsubsection{Experiment Setups}
The training data is used to select the best hyperparameter $a$ with grid search. The search space of hyperparameter $a$ in this work ranges from 0 to 5, incrementing in steps of 0.25.
Two evaluation metrics are used in this experiment: 
\begin{itemize}
    \item Accuracy: measures if state-of-the-art text embeddings can correctly identify semantically similar text $sentence\_2_i$ from negated text $neg\_sentence\_1_i$ which has similar surface form as $sentence\_1_i$.
    \item Correlation: measures if the similarity score of state-of-the-art text embeddings correlates well with human annotated similarity values using Pearson correlation following the evaluation protocol of STSB.
\end{itemize}

\subsubsection{Experimental results}
\begin{table}[t!]
\caption{Performances of text embeddings on STSB dataset: corr is the correlation between original embedding similarity and the human annotations, $corr\_W_{stsb}$  is the correlation between weighted embedding similarity and the human annotations; acc is the accuracy of original embedding;  $acc\_W_{stsb}$ is the accuracy of weighted embedding.}\label{t1}
\begin{adjustbox}{width=0.49\textwidth}
 \centering
\begin{tabular}{lrrrr}
\hline
models & corr & $corr\_W_{stsb}$ & acc & $acc\_W_{stsb}$ \\
\hline
gte-base-en-v1.5 & 87.29 & 86.35 (-0.94) & 79.27 & 80.18 (+0.91)\\
gte-large-en-v1.5 & 84.49 & 83.41 (-1.08) & 74.55 & 79.64 (+5.09)\\
bge-base-en-v1.5 & 86.80 & 85.95 (-0.85) & 75.09 & 79.82 (+4.73)\\
bge-large-en-v1.5 & 87.72 & 86.67 (-1.05) & 79.82 & 86.73 (+6.91)\\
e5-base-v2 & 86.63 & 85.56 (-1.07) & 72.55  & 78.00 (+5.45)\\
e5-large-v2 & 86.26 & 85.58 (-0.68) & 76.73 & 79.27 (+2.54)\\
multilingual-e5 & 85.26 & 84.06 (-1.20) & 76.36 & 82.55 (+6.19)\\
mxbai-embed-large-v1 &\textbf{88.57}  & \textbf{87.70} (-0.87) & \textbf{83.64} & \textbf{89.27} (+5.63)\\
\hline
\textbf{Average (Universal)} & 86.63& 85.66 (-0.97)& 77.25& 81.93 (+4.68) \\
\hline
all-mpnet-base-v2 & 86.37 & 85.48 (-0.89) & 71.45 & 74.18 (+2.73)\\
all-roBERTa-large-v1 & 86.82 & 85.58 (-1.24) & 73.82 & 78.73 (+4.91)\\
\hline
\textbf{Average (Contextual)} & 86.60 & 85.53 (-1.07) & 72.63& 76.46  (+3.83)\\
\hline
\end{tabular}
\end{adjustbox}
\end{table}

The experimental results of state-of-the-art text embeddings on STSB dataset are shown in Table \ref{t1}:  acc is the test accuracy of original embedding;  $acc\_W_{stsb}$ is the accuracy of weighted embedding where the weight is calculated following Equation \ref{wi}. Generally speaking, universal text embeddings have better performance than the two baseline contextual text embedding models.  The proposed dimension weighting method improves the accuracy across all text embeddings with an average improvement of 4.51\%. bge-large-en-v1.5 benefits the most from the proposed method with an improvement of 6.91\% while  gte-base-en-v1.5 has the smallest improvement of 0.91\%.  To study if improved negation awareness would hurt the topic information understanding,  the correlation between human annotated similarity values (mainly based on topic information) and embedding semantic similarity (before and after applying weights) is also calculated as shown in Table \ref{t1}:
corr is the correlation between original embedding similarity and the human annotations, $corr\_W_{stsb}$  is the correlation between weighted embedding similarity and the human annotations. On average, the semantic similarity based on weighted embeddings have slightly reduced correlation (1\% less) with human annotations compared to the original embedding.

In general, universal embeddings benefit more from the proposed solution (+4.68 improvement) than contextual embeddings (+3.83 improvement). 
In terms of embedding model parameter size: larger text embedding models generally have larger dimensions and better negation awareness than smaller  embedding models, with the exception of GTE series: gte-base-en-v1.5 has better performance on both semantic textual similarity (corr) and negation awareness (acc) than gte-large-en-v1.5.

 \subsection{Generalization ability of weights learnt from STSB to SemAntoNeg (more complex negation)}

\textbf{SemAntoNeg Benchmark:} This benchmark from \cite{vahtola2022not} contains more complex negation forms. It assesses the proficiency of embedding models in accurately depicting phrases that incorporate both antonymy and negation. Through a paraphrase task, the model must identify the correct paraphrase among three candidate expressions, each featuring negated sentences and antonym substitutions. Thus, embedding models must understand that adding or removing a negation, coupled with an antonym substitution, retains the sentence's original semantic meaning.  Unlike STSB, there is no human annotation of semantic similarity on SemAntoNeg. Hence, only accuracy metric is used as in \cite{vahtola2022not}.

\begin{table}[t!]
\caption{The generalization ability of the embedding weights learnt from STSB  on SemAntoNeg Benchmark.}\label{t2}
\centering
\begin{adjustbox}{width=0.35\textwidth}

\begin{tabular}{lrr}
\toprule
models & acc & $acc\_W_{stsb}$ \\
\midrule
gte-base-en-v1.5 & 47.12 & 47.17 (+0.05) \\
gte-large-en-v1.5 & 53.58 & 56.18 (+2.60) \\
bge-base-en-v1.5 & 53.67 & 56.78 (+3.11) \\
bge-large-en-v1.5 & 64.68 & 68.73 (+4.05) \\
e5-base-v2 & 54.60 & 56.41 (+1.81) \\
e5-large-v2 & 60.83 & 60.87 (+0.04) \\
multilingual-e5 & \textbf{68.87} & \textbf{73.14} (+4.27) \\
mxbai-embed-large-v1 & 65.80 & 68.91 (+3.11) \\ \hline
\textbf{Average (Universal)} & 58.64 & 61.02 (+2.38)   \\
\hline
all-mpnet-base-v2 & 31.32 & 34.48 (+3.16) \\
all-roBERTa-large-v1 & 35.32 & 37.17 (+1.85) \\ \hline
\textbf{Average (Contextual)} & 33.32 & 35.83 (+2.51) \\
\hline
\bottomrule
\end{tabular}
\end{adjustbox}
\end{table}

Compared to STSB data with simple negation, SemAntoNeg Benchmark is more difficult as it contains negation, antonym and double negation (negation plus antonym). In this section, we evaluate the generalization capability of the simple negation awareness embedding weights, learned from STSB, on the SemAntoNeg Benchmark. The results are presented in Table \ref{t2}. 
The overall performance of universal text embeddings (with the mean performance of 58.64\%) are much better than that of contextual text embedding baselines (with the mean performance of 33.32\%).  The random guess performance on SemAntoNeg Benchmark is 33.33\%, which indicates that contextual text embeddings is prone to predicting the sentence with the highest lexical overlap while ignoring the negation \cite{vahtola2022not}.
On average (across all text embedding models), the embedding weights learnt from STSB improves the performance of the original embedding by 2.41\%. This suggests that the weights learnt from STSB are able to generalize to SemAntoNeg Benchmark.  However, the performance gains for e5-large-v2 and gte-base-en-v1.5 are less noticeable. In terms of model parameter size, it is observed that larger universal text embeddings demonstrate superior comprehension of complex negation compared to their smaller counterparts.

 \subsection{Generalization ability of proposed  solution on SemAntoNeg Benchmark}

\begin{table}[t!]
\caption{The generalization ability of proposed solution on SemAntoNeg Benchmark: $acc\_W_{anto}^{K}$ means the performance of embedding weights learnt from K training samples (K = 200, 500, 100). The mean
and standard deviations of accuracy evaluated over 10 runs are shown in the table.}\label{t3}
\begin{adjustbox}{width=0.5\textwidth}
\centering
\begin{tabular}{l rrrr}
\toprule
models & acc & $acc\_W_{anto}^{200}$  & $acc\_W_{anto}^{500}$ & $acc\_W_{anto}^{1000}$  \\
\midrule
gte-base-en-v1.5 & $47.96$ & $65.72$ & $67.25$ & $67.13$\\ & $\pm0.49$ & $\pm2.85$ & $\pm0.98$ & $\pm0.6$\\ \hline
gte-large-en-v1.5 & $54.16$ & $65.72$ & $68.62$ & $69.21$\\ & $\pm0.49$ & $\pm5.92$ & $\pm4.3$ & $\pm3.35$\\ \hline
bge-base-en-v1.5 & $53.93$ & $71.32$ & $73.09$ & $73.52$\\ & $\pm0.25$ & $\pm1.58$ & $\pm0.61$ & $\pm0.45$\\ \hline
bge-large-en-v1.5 & $65.35$ & $79.92$ & $\textbf{80.97}$ & $80.86$\\ & $\pm0.41$ & $\pm1.09$ & $\pm0.57$ & $\pm0.35$\\ \hline
e5-base-v2 & $54.58$ & $71.34$ & $71.96$ & $72.1$\\ & $\pm0.35$ & $\pm1.2$ & $\pm1.03$ & $\pm0.79$\\ \hline
e5-large-v2 & $61.5$ & $75.24$ & $76.97$ & $77.01$\\ & $\pm0.57$ & $\pm1.87$ & $\pm1.09$ & $\pm0.8$\\ \hline
multilingual-e5 & $\textbf{69.77}$ & $80.10$ & $80.88$ & $80.66$\\ & $\pm0.54$ & $\pm1.67$ & $\pm0.92$ & $\pm1.14$\\ \hline
mxbai-embed-large-v1 & $66.47$ & $\textbf{80.60}$ & $80.93$ & $\textbf{80.99}$\\ & $\pm0.42$ & $\pm0.74$ & $\pm0.41$ & $\pm0.36$\\ \hline
\textbf{Average (Universal)} &59.22 & 73.74 (+14.52) & 75.08 (+15.86) & 75.18 (+15.96) \\ \hline

all-mpnet-base-v2 & $32.1$ & $45.22$ & $45.33$ & $45.66$\\ & $\pm0.52$ & $\pm0.99$ & $\pm0.52$ & $\pm0.78$\\ \hline
all-roBERTa-large-v1 & $35.54$ & $45.0$ & $46.83$ & $47.3$\\ & $\pm0.45$ & $\pm0.97$ & $\pm0.84$ & $\pm1.14$\\ \hline

\textbf{Average (Contextual)} & 33.82 & 45.11 (+11.29) & 46.08 (+12.26)& 46.48 (+12.66)\\ \hline
\end{tabular}
\end{adjustbox}
\end{table}

In this section, the objective is to study whether the proposed embedding  re-weighting solution is able to learn from complex negation (negation + antonym) from  SemAntoNeg Benchmark. Unlike STSB, SemAntoNeg Benchmark has no predefined train test split.    Hence, a stratified repeated random sampling approach is used to achieve a robust estimate of the performance. The random splitting is repeated 10 times, with 32\% as training data (1000x4 samples) and 68\% as test data (2152x4 samples). To test the data efficiency of the proposed solution, we test 3 different training data sizes: 200, 500 and 1000. The corresponding test performances (mean
and standard deviations  of accuracy evaluated over 10 runs) are shown in Table \ref{t3} as $acc\_W_{anto}^{200}$, $acc\_W_{anto}^{500}$, and  $acc\_W_{anto}^{1000}$ respectively. 

\begin{figure}[h]
  \centering
  \includegraphics[width=1\linewidth]{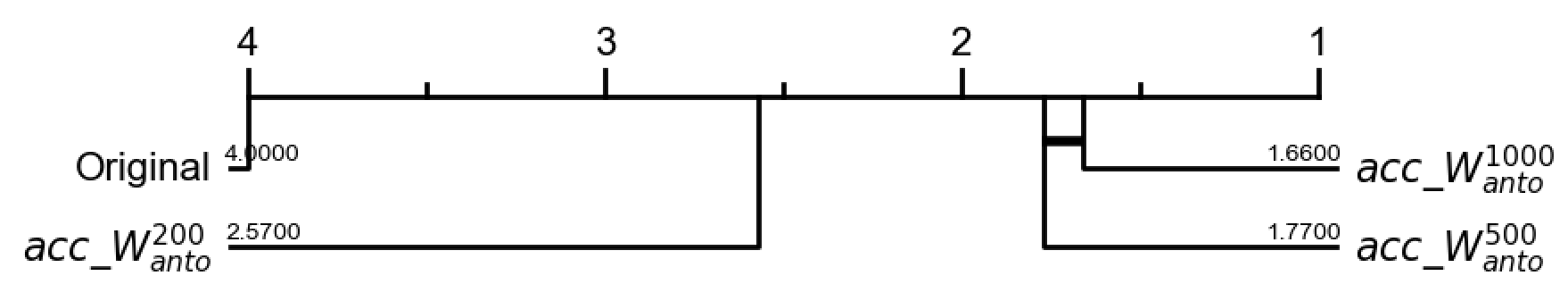}
  \caption{ The CD diagram based on the Wilcoxon-Holm post hoc test
result when alpha is 0.00001. Methods connected with bold black line have
no significant difference.} 
  \label{fig:cd}
\end{figure}

Generally speaking, the proposed weighting method is able to improve the performance on complex negation tasks across all text embedding models with small size training data. 
When the training data size is only 200 ($acc\_W_{anto}^{200}$ in Table \ref{t3}),  the universal text embeddings' performance are improved by 14.52\% on average, while the contextual text  embeddings' performance are improved by 11.29\%.  When increasing the training data size from 200 to 500, the average universal text embeddings' performance are further improved slightly by 1.34\% with reduced standard deviations, while the average contextual text  embeddings' performance are  further improved by 0.97\%. Wilcoxon-Holm post hoc test \cite{cao2019randompr, cao2019random} with Critical Differences (CD) is also done to have an overall statistical comparison as shown in  Figure \ref{fig:cd}: the proposed solution with three training sizes are all significantly better than the original embeddings, but  there is no significant difference between the performance on 500 training datasize and 1000 training datasize (when alpha is 0.00001). Similar to the observations in previous sections: 1. larger universal text embeddings demonstrate superior comprehension of negation compared to their smaller counterparts, 2. universal embeddings benefit more from the proposed
solution than contextual embeddings.  

\subsection{Generalization to Large Language Model based universal text embeddings}
The previous sections have shown that the proposed dimension re-weighting method is able to efficiently improve the negation awareness  of  both universal and contextual text embeddings models. 
Compared to  BERT-based universal embeddings tested in previous sections,  LLM-based universal text embeddings use LLMs as backbones. These backbones  are mostly decoder only models using  causal attention mechanism  which have larger parameter size (mostly 7B) \cite{cao2024recent}. The embedding dimensionality of most  BERT-based universal and contextual embedding models is around 1000 while the embedding dimensionality of  most LLM-based universal text embeddings  is around 4000. 
However, LLM-based text embeddings  have the advantage of  good instruction following ability to deal with potentially mutually contradicted tasks and improve the generalization ability across different tasks, by adding a task specific instruction which describes the nature of the task (e.g. search relevant passages for the query) to the query side. In this case, the same model can provide different task-specific embeddings for different tasks (similar to multilingual-e5) \cite{cao2024recent}.

The objective in this section is to study whether LLM-based  high dimensional task-specific embeddings can be further improved by the proposed dimension weighting method. Three state-of-the-art LLM-based text embeddings are selected: SFR-Embedding-2\_R with Mistral-7B \cite{jiang2023mistral} as backbone, gte-Qwen2-7B-instruct with Qwen2-7B \cite{yang2024qwen2} as backbone and gte-Qwen2-1.5B-instruct with Qwen2-1.5B  \cite{yang2024qwen2} as backbone. The experimental results are shown in Table \ref{t4}: Similar to the observations from previous sections, the proposed data efficient solution can improve significantly (around 12\% improvement on average) the negation awareness of LLM-based text embeddings with small  training data size. 

On average, LLM-based universal text embeddings have better negation awareness than BERT-based universal text embeddings with around 6\% improvement on SemAntoNeg Benchmark. However, this is primarily due to the good performance of gte-Qwen2-7B-instruct, while gte-Qwen2-1.5B-instruct and SFR-Embedding-2\_R are not better than the average performance of BERT-based universal text embeddings with smaller parameter sizes.  Among 4 embeddings with task-specific instructions (3 LLM-based embeddings and multilingual-e5), only multilingual-e5 and gte-Qwen2-7B-instruct show superior comprehension of complex negation compared to their similar-size counterparts. This indicates that different instruction-based embedding models have different instruction-following and instruction-generalization abilities.  On the other side, even though gte-Qwen2-7B-instruct has good performance on SemAntoNeg (81.78\%), the proposed embedding re-weighting method is still able to improve its performance by around 6\%.

\begin{table}[h!]
\caption{The performance of proposed solution with LLM-based embedding models on SemAntoNeg Benchmark: $acc\_W_{anto}^{K}$ means the performance of embedding weights learnt from K training samples (K = 200, 500, 100). The mean
and standard deviations of accuracy evaluated over 10 runs are shown in the table.}\label{t4}
\begin{adjustbox}{width=0.49\textwidth}
\centering
\begin{tabular}{l rrrr}
\toprule
models & acc & $acc\_W_{anto}^{200}$  & $acc\_W_{anto}^{500}$ & $acc\_W_{anto}^{1000}$  \\
\midrule
SFR-Embedding-2\_R & $58.52$ & $71.52$ & $71.32$ & $71.32$\\ & $\pm0.52$ & $\pm1.7$ & $\pm0.85$ & $\pm1.06$\\ \hline
gte-Qwen2-7B-instruct & $\textbf{ 81.78}$ & $\textbf{ 87.43}$ & $\textbf{ 87.52}$ & $\textbf{ 87.44}$\\ & $\pm0.51$ & $\pm0.36$ & $\pm0.38$ & $\pm0.46$\\ \hline
gte-Qwen2-1.5B-instruct & $55.03$ & $72.0$ & $72.24$ & $72.26$\\ & $\pm0.52$ & $\pm0.85$ & $\pm0.47$ & $\pm0.47$\\ \hline
\textbf{Average} & 65.11 & 76.98 (+11.87) & 77.03 (+11.92)& 77.01 (+11.90)\\ \hline
\end{tabular}
\end{adjustbox}
\end{table}

\subsection{Summary}

In this section,  STSB dataset with simple negation is used firstly to test the effectiveness of the proposed dimension re-weighting method. The proposed solution is able to improve the accuracy of negation awareness task of both BERT-based universal text embeddings (4.68\% improvement) and contextual text embeddings (3.83\% improvement). The weights learnt on STSB are subsequently applied to SemAntoNeg Benchmark with complex negation (negation + antonym). The experimental results show that the STSB learnt weights can also improve the complex negation awareness with over 2\% improvement on  BERT-based universal text embedding models. 
Furthermore, the proposed solution is used to learn weights from small training data on SemAntoNeg Benchmark (200, 500 and 1000): performances based on all three training datasizes  are significantly better than the original embedding (with over 10\% improvement) across all  BERT-based embedding models, which shows that the proposed solution is robust and data efficient.   
Finally, we assess the method's  effectiveness on three LLM-based embeddings and show that the proposed method can also improve  the negation-awareness of high dimensional task-specific LLM-based  embeddings. 

\textbf{Sustainability:} The environmental impact of AI solutions has attracted a lot of attention in recent years \cite{bolon2024review, cao2023towards}. Compared to existing solutions, the proposed solution is more lightweight and sustainable.  Specifically, our solution requires a  small training dataset size (200 to 500 instances), necessitates minimal training time (typically under 5 minutes with a single CPU), and incurs significantly lower computational costs, as it operates effectively without the need for GPU resources. Additionally, since our solution does not modify the model parameters, it eliminates the need for storing fine-tuned versions of the model, thereby further reducing resource consumption.

\section{Conclusion}
Recent literature has witnessed the fast progress and development of universal text embeddings. However, due to the bias in popular evaluation benchmarks, the negation awareness capacity of these models remains unclear. In this work, we start with a holistic analysis that exposes a critical “negation blindness” in state‐of‐the‐art universal embeddings. To efficiently bridge this universality gap,  the negation adapter, a lightweight, data‐efficient re‐weighting mechanism that restores negation sensitivity is proposed without modifying the model parameters. The proposed solution is able to improve text embedding models' negation awareness significantly on both STSB with simple negations and SemAntoNeg with complex negations. Furthermore, the proposed solution can also improve the negation awareness of LLM-based task-specific high dimensional embeddings.  An additional benefit of the proposed solution is its flexibility: post-hoc re-weighting is applied exclusively to tasks that need strong negation awareness, while other tasks like topic clustering/classification use the original unweighted embedding, which ensures the universality of universal text embeddings across different tasks.
Future research aims to enhance the proposed semantic adapter for broader applications, integrating the data-efficient solution with automatic training data generation by LLMs to improve the generalization capabilities of universal text embedding models on novel tasks.

\bibliography{sample-base}

\end{document}